\documentclass[conference]{IEEEtran}
\IEEEoverridecommandlockouts
\usepackage{cite}
\usepackage{amsmath,amssymb,amsfonts}
\usepackage{algorithmic}
\usepackage{graphicx}
\usepackage{textcomp}
\usepackage{xcolor}
\usepackage{booktabs} 
\def\BibTeX{{\rm B\kern-.05em{\sc i\kern-.025em b}\kern-.08em
    T\kern-.1667em\lower.7ex\hbox{E}\kern-.125emX}}
\usepackage{multirow}
\usepackage[utf8]{inputenc}
\usepackage[table]{xcolor} 
\usepackage{caption}
\usepackage{booktabs} 
\usepackage{float}
\usepackage{cite}

\definecolor{LightBlue}{rgb}{0.88, 0.94, 1}
\usepackage{subcaption} 
\usepackage[compatibility=false]{caption} 
\usepackage{fancyhdr}
\usepackage{lipsum}
\usepackage{subcaption} % Modern and preferred over subfigure

\fancypagestyle{firstpage}{
  \fancyhead[L]{\rule{0pt}{0pt}
} 
  \fancyfoot[L]{\rule{0pt}{0pt} }
}

\title{AI-Powered Deepfake Detection Using CNN and Vision Transformer Architectures}
\author{
    Md Sifatullah Sheikh\textsuperscript{1}, 
    Urmi Kirtonia\textsuperscript{1}, 
    Nuzath Tabassum Arthi\textsuperscript{1}, 
    Md Al-Imran\textsuperscript{1*}%
    \\
    \textsuperscript{1}Department of Computer Science and Engineering, East West University,\\
    A/2 Aftabnagar, Dhaka 1212, Bangladesh\\
    Emails: \{mdsifatullahsheikh, urmikirtonia32, tabassumarthi2001\}@gmail.com, al.imran@ewubd.edu
    
}

 \usepackage[pscoord]{eso-pic}
\newcommand{\placetextbox}[3]{
 \setbox0=\hbox{#3}
 \AddToShipoutPictureFG*{ \put(\LenToUnit{#1\paperwidth},\LenToUnit{#2\paperheight}){\vtop{{\null}\makebox[0pt][c]{#3}}}
 }
 }
\placetextbox{.23}{0.055}{\small{ 979-8-3315-9474-9/25/\$31.00~\copyright 2025 IEEE}}

\begin{document}
\maketitle

\thispagestyle{firstpage}
\begin{abstract}
The increasing use of artificial intelligence-generated deepfakes creates major challenges in maintaining digital authenticity. Four AI-based models, consisting of three CNNs and one Vision Transformer, were evaluated using large face image datasets. Data preprocessing and augmentation techniques improved model performance across different scenarios. VFDNET demonstrated superior accuracy with MobileNetV3, showing
efficient performance, thereby demonstrating AI’s capabilities for dependable deepfake detection.
\end{abstract}

\renewcommand{\thefootnote}{\fnsymbol{footnote}}
\footnotetext[1]{Corresponding author}

\section{Introduction}
The development of deepfakes altered the landscape of digital media, whereby it enabled the creation of hyper-realistic content. There have been reported cases where creative applications were lauded, whereas the lack of unethical use has permeated. 2017 \cite{nguyen2023deepfake} saw widespread misuse of such technology with fake celebrity content and non-consensual altered porn, igniting grave concerns about AI and personal privacy issues.\par
\vspace{1mm} A 2020 report from Sensity AI (formerly Deeptrace) \cite{sensity2021state} reported that  96\% of online deepfake videos were pornographic and created without consent, which mainly targeted women before being distributed through adult websites and the dark web. The inappropriate use of deepfake technology leads to both psychological trauma and reputational harm while resulting in decreased public confidence in digital content. The use of deepfakes extends beyond privacy threats because they present significant dangers to democracy and security by supporting political manipulation along with identity theft and scams, and widespread disinformation. Fake media \cite{chesney2019deep} spread without control endangers legal systems because it makes digital evidence validation more difficult to perform, according to Chesney and Citron. \par
\vspace{1mm} To counter the growing threat presented by more realistic deepfakes, this study suggests a machine learning and deep learning-based method to recognize modified face photos. We specifically examine the efficiency of the Deepfake Convolutional Network (DFCNET), MobileNetV3, ResNet50, and Vision Fake Detection Network (VFDNET) in discriminating between real and modified information. This presents the key research question: \textbf{Which of these models performs the most accurately and strongly in detecting deepfake images using various alteration techniques?} Our goal is to contribute to the development of accurate detection systems that will help restore public trust in digital media.

\section{Literature review}
Deepfake detection research is the main theme of this part, which focuses on the performance of models trained on specific datasets. \par
\vspace{1mm} The dense CNN (D-CNN) developed by Patel et al. (2023)  \cite{patel2023improved} obtained 97.2\% accuracy throughout their real and fake image dataset. The D-CNN outperformed MesoNet variants by achieving 99.33\% on GDWCT  and 99.17\% on StarGAN, while performance reduced to 94.67\% for high-resolution images from StyleGAN2. \par
Vision Transformers (ViT) combined with 40,000 Kaggle images became the tool of choice for Ghita et al. (2024) \cite{10646310}  as they reached 89.91\% accuracy. The approach showed potential but required extensive computational resources and achieved a moderate accuracy level without testing across different datasets. \par
DenseNet-121 and VGG16 models from Alkishri et al. (2023) \cite{article} achieved 99\% detection accuracy when they processed 140K Real and Fake Faces data with a DFT-based frequency analysis method. The detection system experienced reduced accuracy when handling low-resolution images due to insufficient frequency information. \par
The research by  Zhang et al. (2022) \cite{zhang2022heterogeneous} presents an ensemble learning technique that merges gradient and frequency and texture characteristics. Using a training dataset of CelebA and  five GAN-based fakes, the model reached 97.04\%
\section{Methodology}
\begin{figure}[h]
    \centering
    \includegraphics[width=\linewidth]{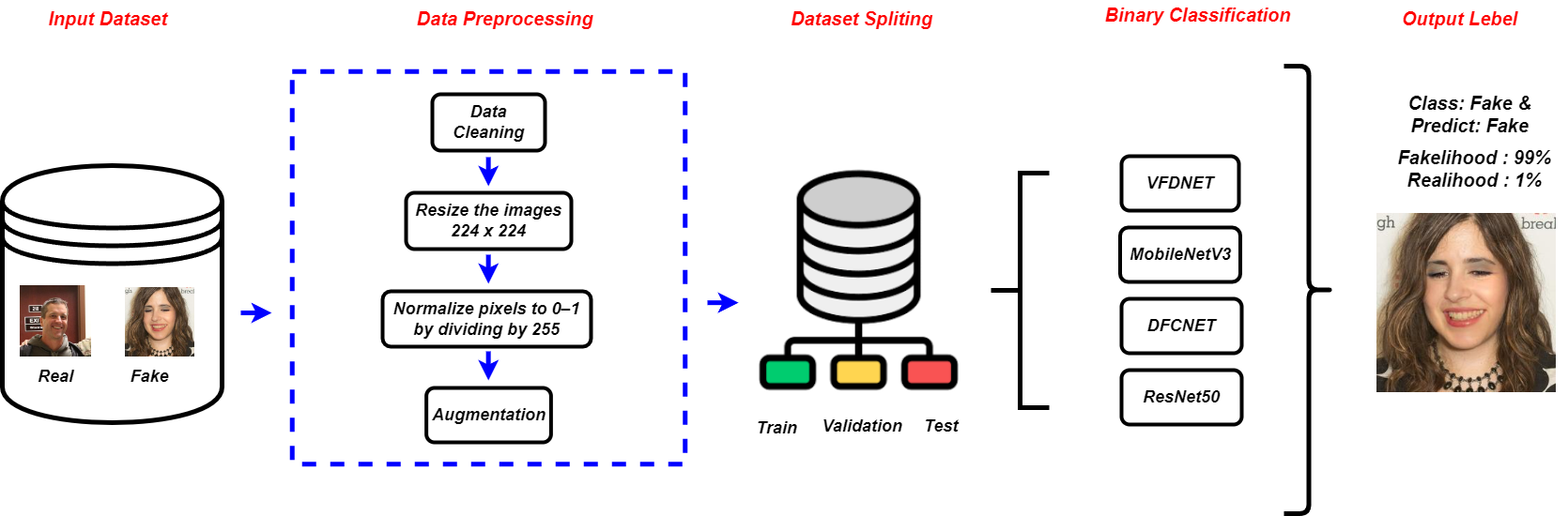}
    \caption{Schematic Workflow of the Proposed Method}
    \label{fig: Workflow Schematic Representation.}
\end{figure}

 For deepfake detection, this study focuses on vision transformers and convolutional neural networks. Major preprocessing processes and feature analyses were carried out, as seen in Figure \ref{fig: Workflow Schematic Representation.}.
\subsection{Preprocessing Pipeline for Enhanced Model Generalization}
Data sets require specific preprocessing steps such as normalizing pixel values $[0,1]$ and resizing images $224 \times 224$ to maximize model performance. The preprocessing procedure converts RGB images to grayscale batches of sixteen. Real and fake data are classified as binary under labels 0 and 1. Data generalization improves through image augmentation techniques, including rotation, scaling, resizing, and flipping, which also decrease overfitting. Implementing histogram analysis guarantees consistent data distributions between training, validation, and test data sets.
\subsection{Fundamental CNN Layers}
Convolutional Neural Networks (CNNs) are composed of essential layers that extract, reduce, and categorize input data. This section describes the basic layers that make up the foundation of CNN architectures.
\subsubsection{Pooling Layer} 
The pooling layer reduces the size of the image while keeping the key features. The output is called a pooled feature map. The inclusion of these layers is important because not only do they help reduce overfitting, but they also reduce the cost of computation.
\subsubsection{Convolutional Layer} 
Convolutional layer activations can be expressed as a tensor of size $Height$ $X$ $Width$ $X$ $F_m$, where $F_m$ represents the number of feature maps. The algorithm divides \cite{liu2015treasure} the input picture into $Height$ $X$ $Width$ sections and uses D-dimensional feature maps to identify visual patterns from each sector. Each filter performs a convolution on the input images to discover a particular feature and adds a bias term, as represented in Equation \eqref{eq:A graph neural network.}. \par
\begin{equation}
h^{k+1} = \sigma \left( W^k \otimes h^k + b^k \right)
\label{eq:A graph neural network.}
\end{equation}
The convolution operation \cite{alsaleh2023space}  in this formula $\otimes$ is denoted by $h^k$ for input feature map k and by $W^k$ for filter weight matrix and by $b^k$ for bias term and by $h^{k+1}$ for output feature map together with $\sigma$ for the activation function. 
\subsubsection{Fully Connected (FC) or Dense Layer}
\begin{figure}[h]
    \centering    \includegraphics[width=1\linewidth,height=5cm]{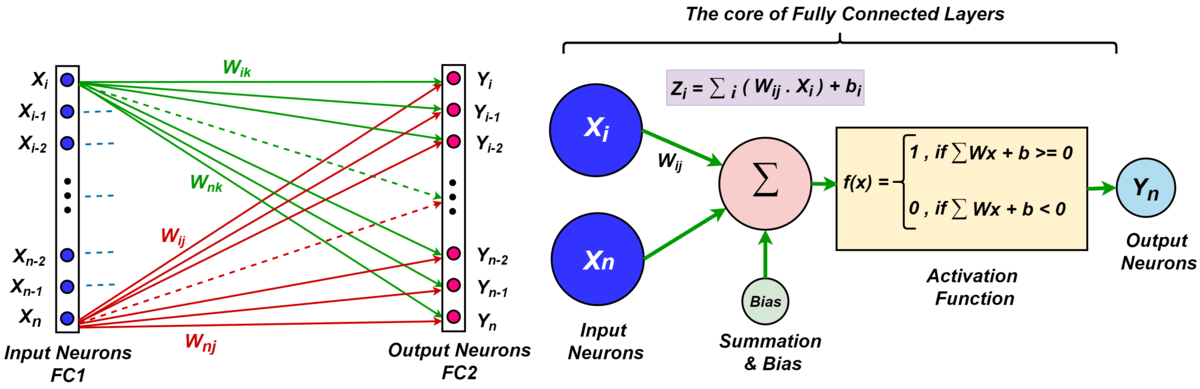}
    \caption{Fully Connected (FC) layer with two layers.}
    \label{fig: Fully Connected (FC) layer with two layers.}
\end{figure}

A dense layer within a neural network is a layer where every neuron in the present layer is connected to every neuron in the previous layer, thus, every input node is connected to every output node. Figure \ref{fig: Fully Connected (FC) layer with two layers.} represents a network that has two FC layers, where input and output neurons are shown. The two layers are indicated as Fully Connected 1 as FC1 and Fully Connected 2 as FC2. Suppose that x is the output of FC1 where $x \in \mathbb{R}^{N_1 \times 1}$. Assume that W represents the weight matrix of FC2 where $W \in \mathbb{R}^{N_1 \times N_2}$ and $w_i$ is the $i\text{-th}$ column vector of W. The weight vector of the corresponding neuron $i\text{-th}$ in the FC2 layer is represented by each column $w_i$. Therefore, the output of FC2 \cite{ma2017equivalence} is given by $W^T$x.
\subsubsection{Flatten Layer} 
The neural network uses a flattening layer to transform the output data produced by the previous network segment into a one-dimensional matrix. The flattening layer can filter individual neurons. The flattened layer decreases the number of weights that exist between the convolutional and fully connected layers.
\subsection{Commonly used Activation Functions}
In neural networks, each neuron in the same layer has the same activation function. The model is trained by engaging gradient descent and backpropagating error signals through each neuron. 
\subsubsection{Rectified Linear Unit }
A Convolutional Neural Network (CNN) implements ReLU activation functions to simplify operations within its neural network layers. The use of leaky ReLU activation functions \cite{xu2015empirical} helps the network process negative values more efficiently, as depicted by Equation \eqref{eq: max input for the neuron}.
\begin{equation}
\label{eq: max input for the neuron}
\mathit{ReLU}(z) = \max(0, \mathit{z})
\end{equation}
The function represents z as the neuron input. When z is above 0, the function delivers z  back. The function generates an output of 0 whenever z remains equal to or lower than 0. The mathematical  representation of the ReLU and Leaky ReLU functions uses the following equations \eqref{eq: ReLU equation.} and \eqref{eq: LeakyReLU equation.}:
\begin{equation}
\text{ReLU}(z) = 
\begin{cases}
z, & \text{if } z > 0 \\
0, & \text{if } z \leq 0
\end{cases}
\label{eq: ReLU equation.}
\end{equation}
\begin{equation}
\text{Leaky-ReLU}(z) = 
\begin{cases}
z, & \text{when } z > 0 \\
0, & \text{when } z \leq 0
\end{cases}
\label{eq: LeakyReLU equation.}
\end{equation}
\subsubsection{Sigmoid Activation Function}
The S-shaped curve of the sigmoid activation function ranges between one and zero. It aids neural networks in learning from the data and is used for binary classification, but it is highly sensitive to input perturbations. Equation \eqref{eq:sigmoidactivationfunction1} and \eqref{eq:sigmoidactivationfunction2} denote that Sigmoid Activation function $\mu$(z) and Hyperbolic Tangent function $\tau(z)$ can be expressed as:
\begin{equation}
    \mu(z) = \frac{1}{1+exp^{-z}}
    \label{eq:sigmoidactivationfunction1}
\end{equation} 
\begin{equation}
    \tau(z) = \frac{\exp^z - \exp^{-z}}{\exp ^z + \exp^{-z}}
    \label{eq:sigmoidactivationfunction2}
\end{equation}
Where $exp^z$ means the exponential function, which grows swiftly as z increases, and $exp^{-z}$ means the same function but mirrored, meaning it decreases as z increases.
\subsection{Framework for the detection of fake images}
Four models, including three CNN-based and one transformer-based, were fine tuned to classify real and fake deepfake images.
\subsubsection{\textbf{Deepfake Convolutional Network (DFCNET)}}
The DFCNET model is an unknown function that uses various layers connected one after another. For each layer, we create a feature map \( z_i \in \mathbb{R}^{m_i \times n_i \times c_i} \), which is obtained from the output of the previous layer \( z_{i-1} \). The input image $x$ serves as the first layer \(z_{0}\), and the final layer generates the output image $y$.  \par
\vspace{1mm} Each input channel is convolved with a specific filter, summed pixel wise with a bias, and passed through a non linear function. Repeating this with various filters produces multiple output channels. The output \( a_i^j \) \cite{pelt2018mixed} of channel $j$ is given in Equations \eqref{eq:activation_equation} and \eqref{eq:filtering_equation}:
\begin{equation}
    a_i^j = \phi\left( f_{ij}(a_{i-1}) + \beta_{ij} \right)
\label{eq:activation_equation}
\end{equation}
\begin{equation}
    f_{ij}(a_{i-1}) = \sum_{k=0}^{c_{i-1}} \mathcal{F}_{\kappa_{ijk}}(a_{i-1}^{k})
\label{eq:filtering_equation}
\end{equation}
Here, $\phi$ is a nonlinear activation function. $\beta_{ij} \in \mathbb{R}$ is the bias and $f_{ij}$ convolves each channel of $c_{i-1}$ with a filter and sums the results.\par
\vspace{1mm} For incorporating nonlinearity, we built our bespoke CNN with three 2D convolutional layers and max-pooling with ReLU activation. Dropout was introduced after every convolutional layer as a safeguard against overfitting. Then, a deep layer that had 256 dimensions of ReLU-activated units was employed to represent the high level features. This layer utilized a sigmoid classification function to produce the binary classifier output. The training process involved the use of the Adam optimizer, which led to model evaluation based on binary cross-entropy loss and accuracy metrics.
\subsubsection{\textbf{Vision Fake Detection Network (VFDNET)}}
In image classification, the Vision Transformer (ViT), a pioneering transformer-based design, has performed similarly to or better than CNNs. Its ability to collect global contextual relationships through self-attention makes it ideal for tasks like bogus picture identification. In this paper, we also suggest a ViT-based model to solve the classification problem. \par
\begin{figure*}[h]
\centering
\includegraphics[width=\linewidth]{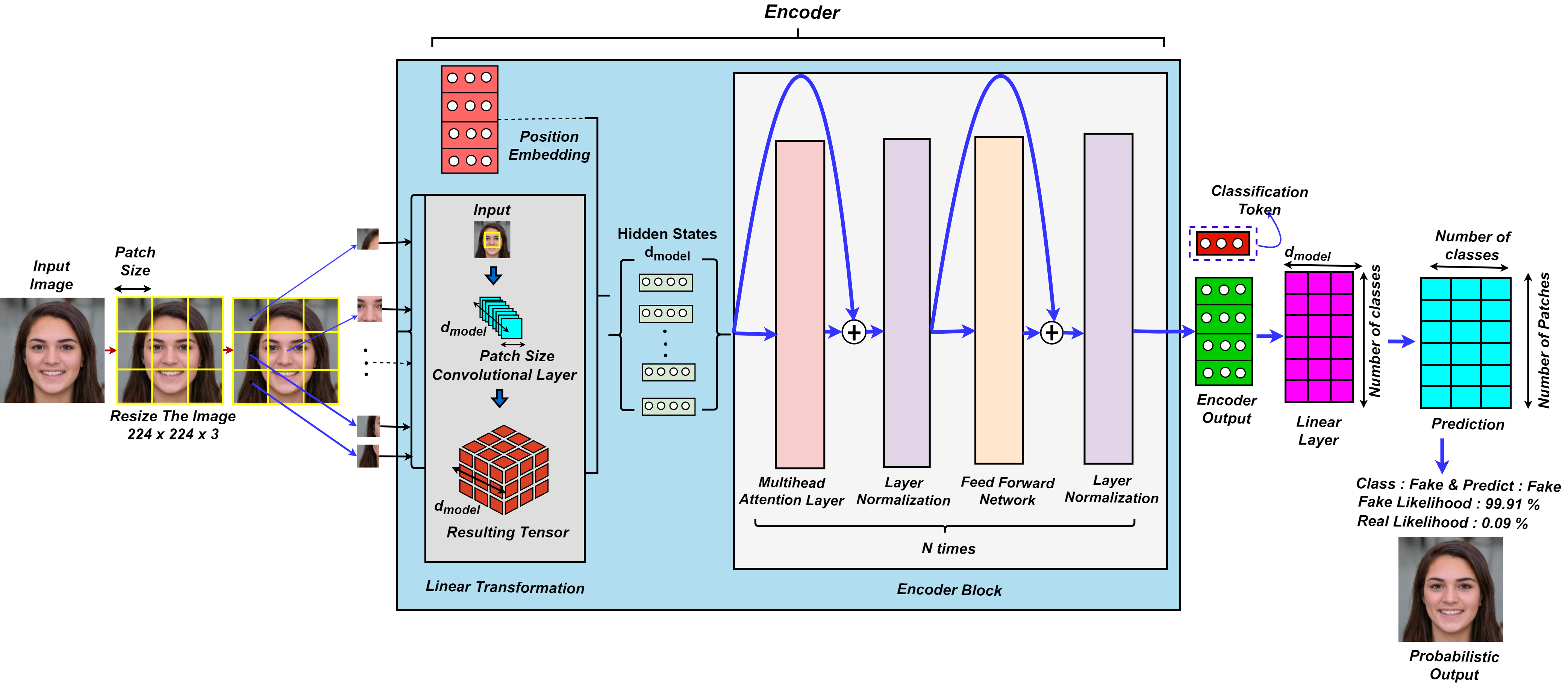}
 \caption{Architecture of the Vision Fake Detection Network (VFDNET)}
\label{fig:Architecture of a ViT.}
\end{figure*}
\vspace{1mm} As illustrated in Figure \ref{fig:Architecture of a ViT.}, the proposed VFDNET model first divides the input image into fixed-size patches, linearly projecting each into tokens to form a sequence. Three data augmentation techniques—AutoAugment\_transform, RandAugment\_transform, and Auto\_RandAugment\_transform—were applied to enhance model robustness. All input images were resized to 224 × 224 × 3. The augmented datasets were organized using the ImageFolder class and loaded via DataLoader for training and evaluation. The Adam optimizer was employed due to its adaptive learning rate capability. The transformer encoder \cite{chen2021crossvit} design uses many blocks that consist of multi-headed self-attention (MSA) combined with a feedforward network (FFN). The feedforward network implements a two-layer multilayer perceptron with the hidden layer expanding at a rate of r, followed by one GELU non-linearity applied to the first linear layer. Every block within the transformer receives layer normalization (LN), which supports residual connections made after each block. The VFDNET input, $h_0$, and the $k$th block transformation process can be formally described in Equation \eqref{eq:transformer_block_modified}:
\begin{equation}
\begin{cases}
\mathbf{h}_0 = [\mathbf{h}_{\text{class}} \, \Vert \, \mathbf{h}_{\text{patch}}] + \mathbf{h}_{\text{position}} \\[10pt]
\mathbf{q}_k = \mathbf{h}_{k-1} + \operatorname{MSA}(\operatorname{LN}(\mathbf{h}_{k-1})) \\[10pt]
\mathbf{h}_k = \mathbf{q}_k + \operatorname{FFN}(\operatorname{LN}(\mathbf{q}_k))
\end{cases}
\label{eq:transformer_block_modified}
\end{equation}
where $\mathbf{h}_{\text{class}} \in \mathbb{R}^{1 \times C}$ and $\mathbf{h}_{\text{patch}} \in \mathbb{R}^{N \times C}$ are the class and patch tokens, respectively, and $\mathbf{h}_{\text{position}} \in \mathbb{R}^{(1+N) \times C}$ is the position embedding. C is the size of the matching embedding dimension, and N is the number of extracted patch tokens.
\subsubsection{\textbf{MobileNetV3}}
The tuning process of MobileNetV3 \cite{howard2019searching} for mobile phone CPUs results from hardware Network Architecture Search (NAS) in combination with the NetAdapt algorithm. MobileNetV3 consists of two implemented models, which are MobileNetV3 Large and MobileNetV3 Small. Each of these models was specifically designed to address either high or low resource usage needs. The models have been customized and used to accomplish both object detection and semantic segmentation tasks. MobileNetV3 enhances its architecture to deliver better performance using minimal computational resources.
\subsubsection{\textbf{ResNet50}}
ResNet50 is a deep neural network architecture that consists of 50 weight layers. ResNet50 is a CNN architecture model that normalizes the concepts of residual learning and skip connections. ResNet50 belongs to the residual network family. The ResNet50 contains 48 convolution layers, which include a single maxpool layer and an average pool layer. This model processes data through an initial sequence of layers, which include low-level feature extraction and ReLU activation through convolution operations and batch normalization, and max pooling.
\section{ Experimental Set}
In this section, we divide the discussion into four parts. The first part is to use the DFCNET, ResNet50, VFDNET, and MobileNetV3 networks to train and test the 140K real and fake faces dataset. 
\subsection{Dataset Description}
The dataset containing deepfake images is accessible through Kaggle from Xhlulu, who uploaded it, and contains 140K real and fake faces. The dataset contains 70000 photos of authentic faces, which come from the Flickr dataset that NVIDIA compiled. An additional 70000 fake face images exist in this dataset, which originates from the 1 Million Fake Faces dataset developed through StyleGAN and distributed by Bojan.
\subsection{Dataset Splitting}
We have achieved pre-processing, exploratory data analysis, and feature selection on the dataset. In feature selection, we feature the intensity of the image pixels of the train, test, and validation datasets. The data set contains a training, test, and validation dataset. The data sampled were divided into 70\% train data, 15\% validation data, and 15\% test data. 
\subsection{Dataset Training and Validation}
For the DFCNET, ResNet50, and MobileNetV3 models, images are rescaled to normalize pixel values to a range of [0,1]. $ImageDataGenerator$(rescale=1./255) is used for normalization. Images are resized to 224 x 224 pixels (target\_size=(224,224)) and data is processed in small batches of 16 images (batch\_size=16). Image data loads from flow\_from\_directory (train\_dir,val\_dir,test\_dir). The class\_mode ='binary' sets the binary labels (real vs fake). For training VFDNET, we used the AutoAugment Pipeline($AutoAugment$, resizing, center cropping, normalization, and tensor conversion) to enhance the training data. We also used the RandAugment Pipeline and the Combined Auto and RandAugment Pipeline.
\subsection{Evaluation Metrics}
Here we outline the evaluation criteria and results analysis. Evaluation of experimental performance in classification studies primarily relies on the accuracy metric, which we express with the mathematical representation of Equation \eqref{eq:acc}:
\begin{equation}
\label{eq:acc}
    \mathit{Accuracy} = \frac{\textit{Correct Predictions}}{\textit{Total Predictions}} = \frac{TP + TN}{TP + TN + FP + FN}
\end{equation}
Where TP denotes true positives, TN denotes true negatives, FP denotes false positives, and FN denotes false negatives. \par
The assessment of image classification system performance uses Precision and Recall as evaluation metrics. Precision calculates the percentage of correct images from the total number of classified images. The mathematical expression represented in Equation \eqref{eq:presion}:
\begin{equation}
\label{eq:presion}
    \mathit{Precision} = \frac{\textit{True Positives}}{\textit{True Positives} + \textit{False Positives}}
\end{equation}
Here, FP represents misclassified images, and TP represents the correctly classified images.\par
The recall calculates the percentage of accurately predicted pictures relative to the dataset size. The
mathematical form in Equation \eqref{eq:recall}:
\begin{equation}
\label{eq:recall}
\mathit{Recall} = \frac{\textit{True Positives}}{\textit{True Positives} + \textit{False Negatives}}
\end{equation}
Here, FN denotes false negatives, the images that fit the correct class but are misclassified by the classifier.\par
The mathematical notation for  F1-score shows it as the weighted average of precision and recall values. The mathematical representation of F1-Score is  given by Equation  \eqref{eq:f1}:
\begin{equation}
\label{eq:f1}
\mathit{F1-Score} = 2 *\frac{\mathit{Precision} *\mathit{Recall}}{\mathit{Precision}+ \mathit{Recall}}
\end{equation}
\section{Results Analysis }
The following subsections present and discuss the experimental results and the comparative analysis.
\subsection{Significance of Performance Across Epochs}
\begin{figure}[htbp]
    \centering
    % Row 1
    \begin{subfigure}[b]{0.24\textwidth}
        \includegraphics[width=\linewidth,height=4.5cm]{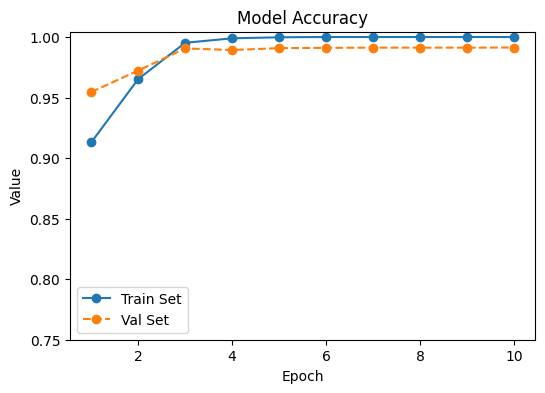}
        \caption{VFDNET Accuracy}
        \label{fig:vit_acc}
    \end{subfigure}
    \hfill
    \begin{subfigure}[b]{0.24\textwidth}
        \includegraphics[width=\linewidth,height=4.5cm]{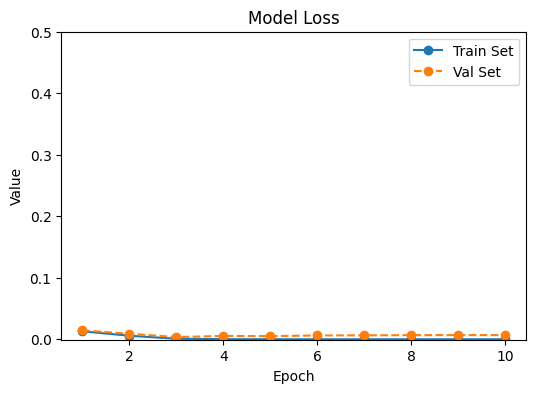}
        \caption{VFDNET Loss}
        \label{fig:vit_loss}
    \end{subfigure}
    \vskip1em
    \begin{subfigure}[b]{0.24\textwidth}
        \includegraphics[width=\linewidth,height=4.5cm]{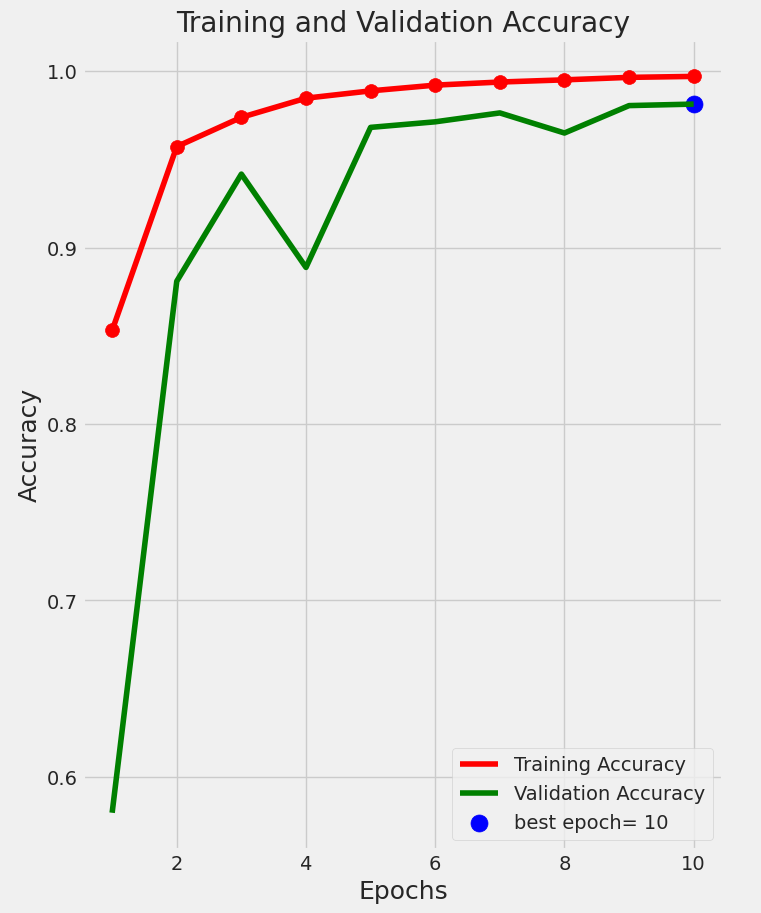}
        \caption{MobileNetV3 Accuracy}
        \label{fig:mobile_acc}
    \end{subfigure}
    \hfill
    \begin{subfigure}[b]{0.24\textwidth}
        \includegraphics[width=\linewidth,height=4.5cm]{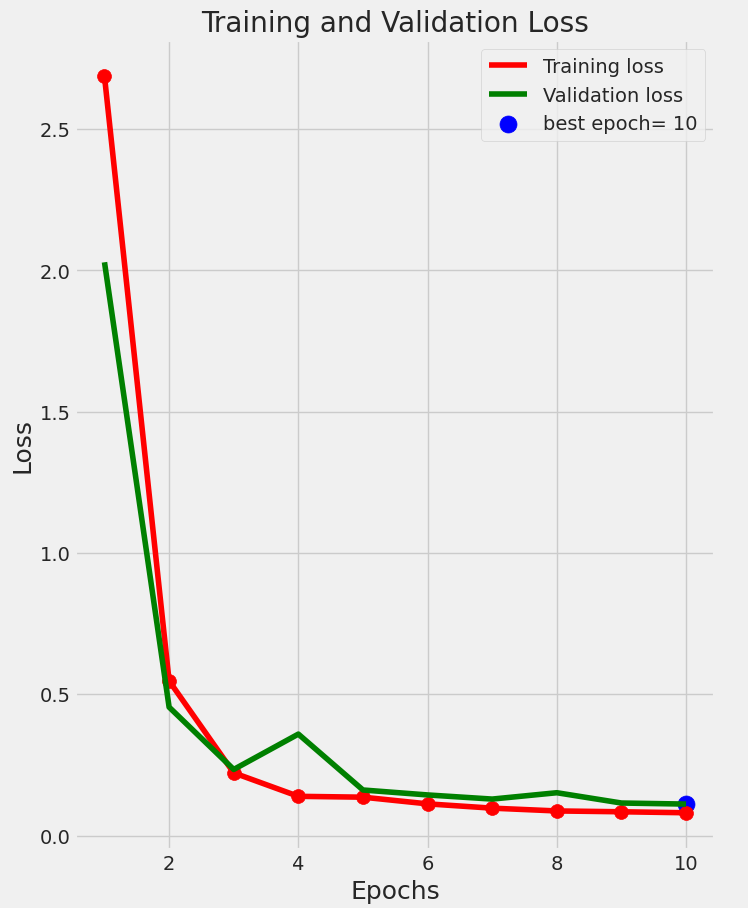}
        \caption{MobileNetV3 Loss}
        \label{fig:mobile_loss}
    \end{subfigure}
    \vskip1em
    \begin{subfigure}[b]{0.24\textwidth}
        \includegraphics[width=\linewidth,height=4.5cm]{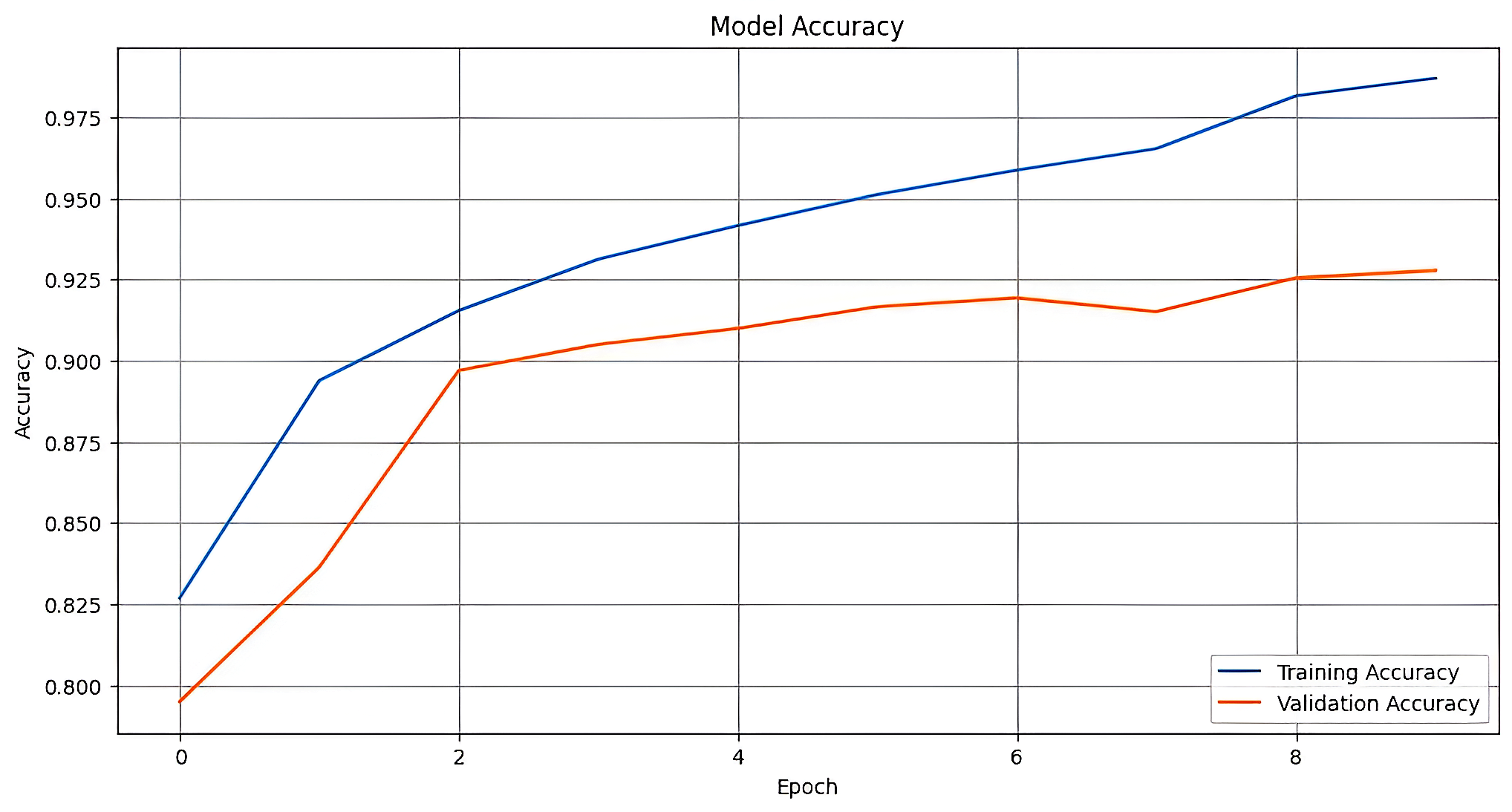}
        \caption{ResNet50 Accuracy}
        \label{fig:resnet_acc}
    \end{subfigure}
    \hfill
    \begin{subfigure}[b]{0.24\textwidth}
        \includegraphics[width=\linewidth,height=4.5cm]{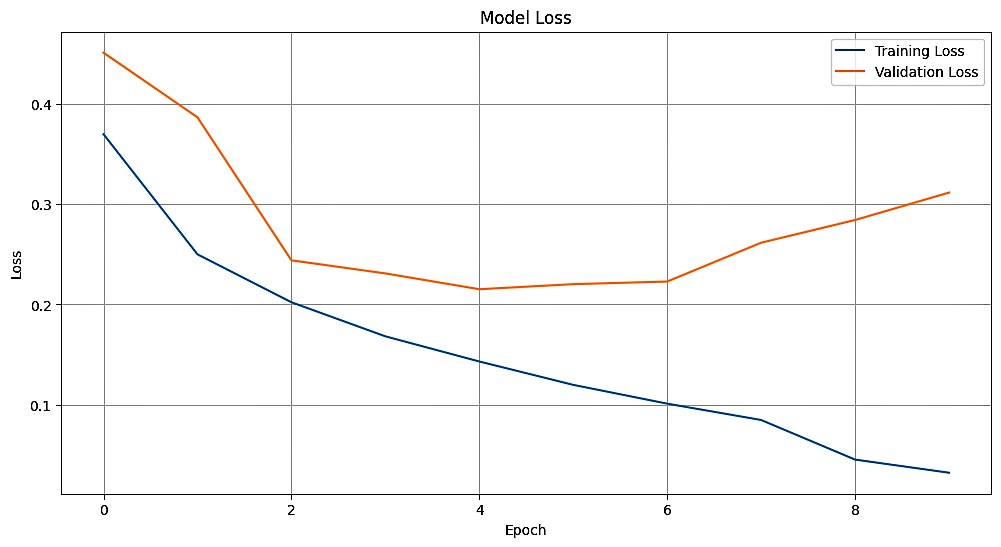}
        \caption{ResNet50 Loss}
        \label{fig:resnet_loss}
    \end{subfigure}
    \vskip1em
    \begin{subfigure}[b]{0.24\textwidth}
        \includegraphics[width=\linewidth,height=4.5cm]{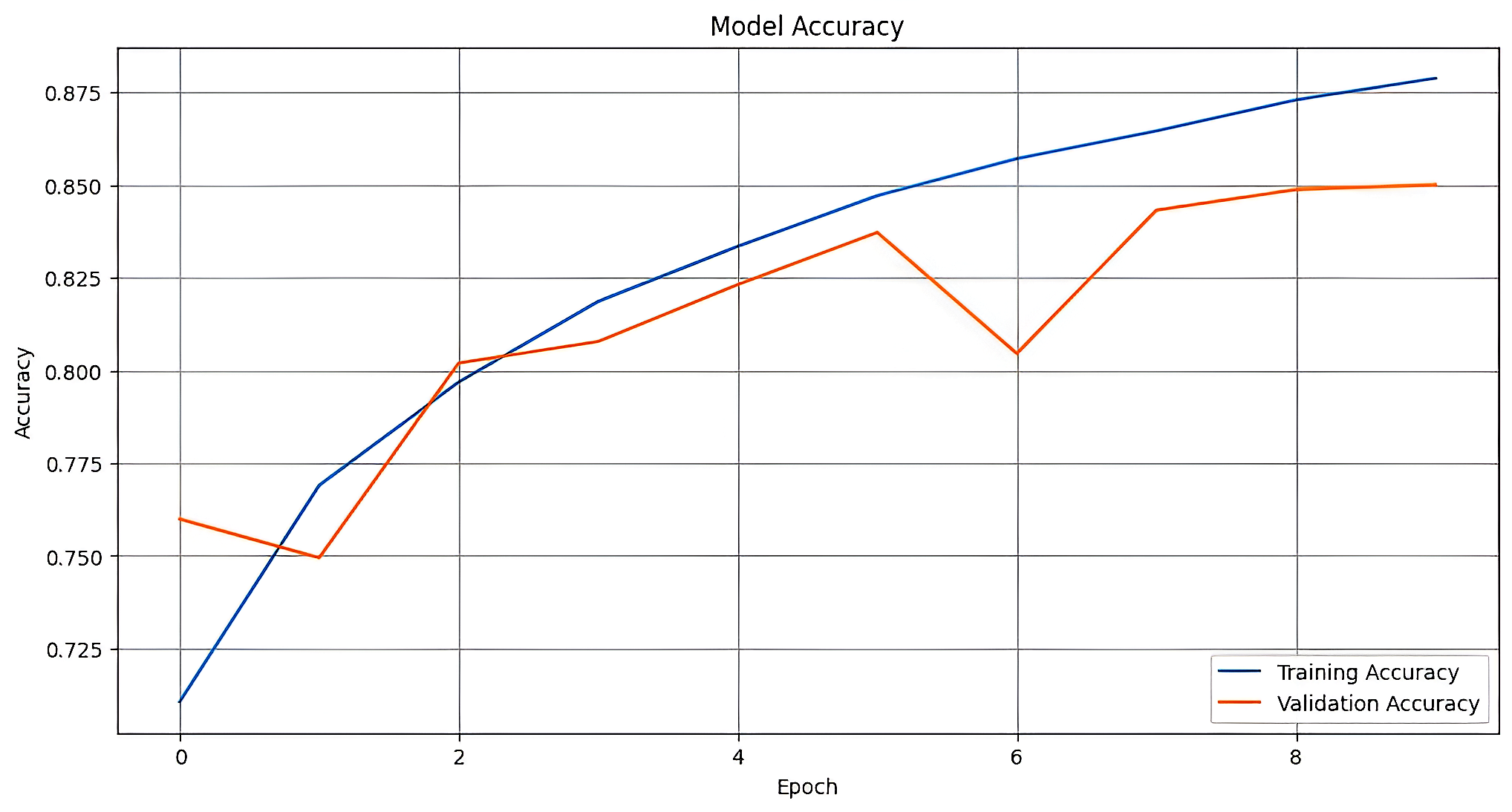}
        \caption{DFCNET Accuracy}
        \label{fig:cnn_acc}
    \end{subfigure}
    \hfill
    \begin{subfigure}[b]{0.24\textwidth}
        \includegraphics[width=\linewidth,height=4.5cm]{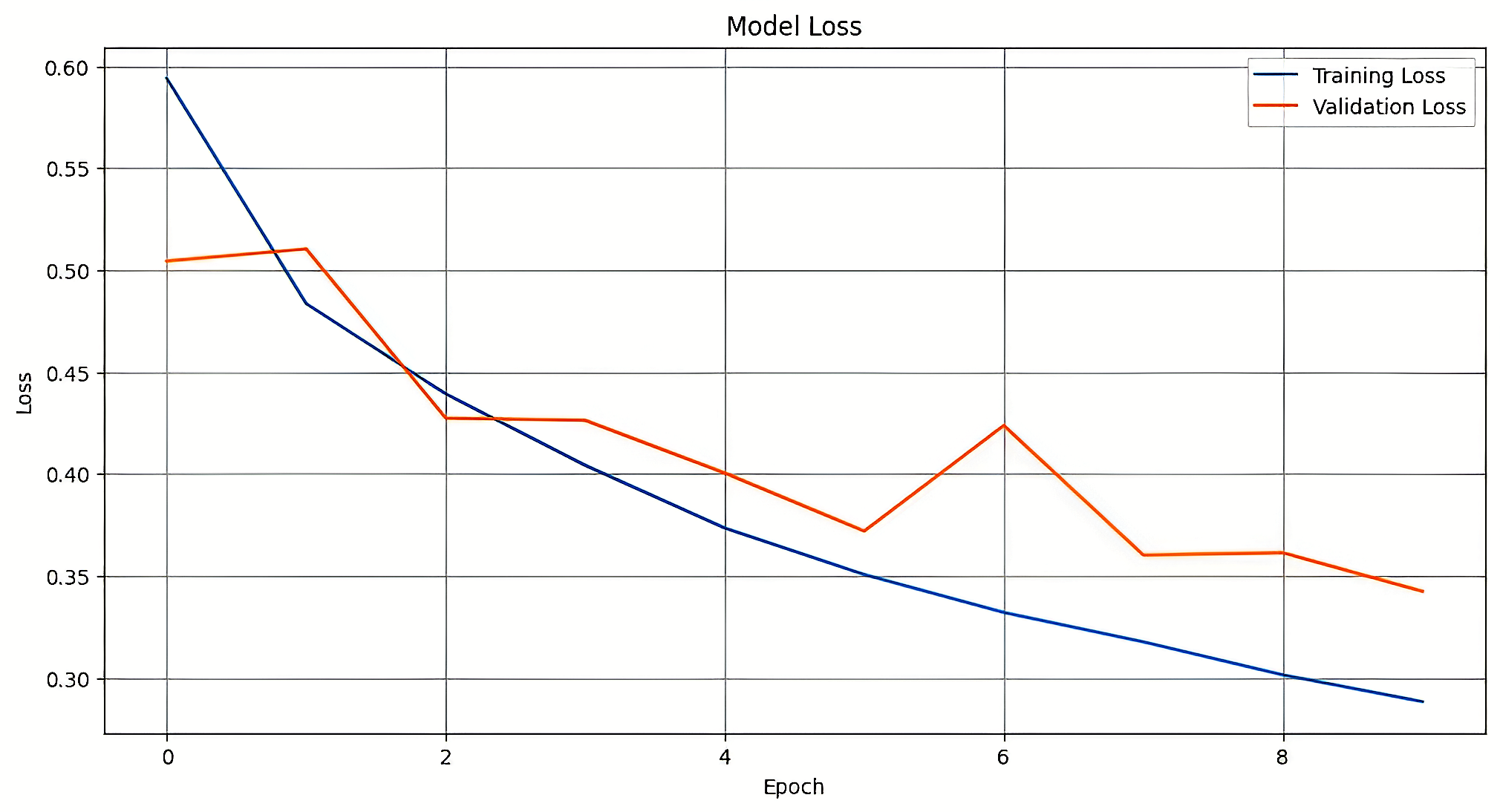}
        \caption{DFCNET Loss}
        \label{fig:cnn_loss}
    \end{subfigure}
    
    \caption{Training and validation accuracy and loss curves for all employed models.}
    \label{fig:all_models_curves}
\end{figure}
Figure \ref{fig:all_models_curves} and Table \ref{tab:accloss} collectively present the training and validation performance of the proposed models. The VFDNET outperformed others, achieving perfect training accuracy and the lowest validation loss of 0.0068, indicating strong generalization and convergence stability. MobileNetV3 also demonstrated competitive performance with a high validation accuracy of 0.9814 and low loss. In contrast, ResNet50 and DFCNET exhibited higher validation losses and noticeable performance gaps, suggesting overfitting. The plotted learning curves further corroborate these findings, visually highlighting the superior learning dynamics of transformer-based and lightweight architectures.
\begin{table}[ht]
    \centering
    \caption{Performance Benchmark Across Models}
    \label{tab:accloss}
    \renewcommand{\arraystretch}{1.3}
    \setlength{\tabcolsep}{4pt} % reduce horizontal padding
    \scriptsize % reduce font size
    \begin{tabular}{lcccc}
        \hline
        \textbf{Proposed} & \textbf{Train Acc.} & \textbf{Train Loss} & \textbf{Val Acc.} & \textbf{Val Loss} \\
        \hline
        VFDNET & 1.0000 & 0.00001 & 0.9913 & 0.0068 \\
        ResNet50 & 0.9870 & 0.0318 & 0.9278 & 0.3117 \\
        MobileNetV3 & 0.9970 & 0.0805 & 0.9814 & 0.1113 \\
        DFCNET & 0.9882 & 0.0325 & 0.9446 & 0.1842 \\
        \hline
    \end{tabular}
\end{table}
\subsection{Results Performance Analysis of Employed Techniques}
Table \ref{tab:model_performance} presents the comparative performance of the proposed and baseline models. The VFDNET model achieved the highest accuracy 99.13\% and demonstrated balanced precision, recall, and F1-score 99.00\%, indicating superior generalization. MobileNetV3 also performed robustly with 98.00\% accuracy. In contrast, DFCNET achieved moderate accuracy 95.76\% but maintained high precision and recall with 92\% and 91\%. ResNet50 recorded the lowest performance across all metrics 84.28\%, underscoring the effectiveness of transformer-based and lightweight architectures over traditional CNNs.\par
\begin{figure}[h]
    \centering
    \includegraphics[width=1\linewidth,height=7cm]{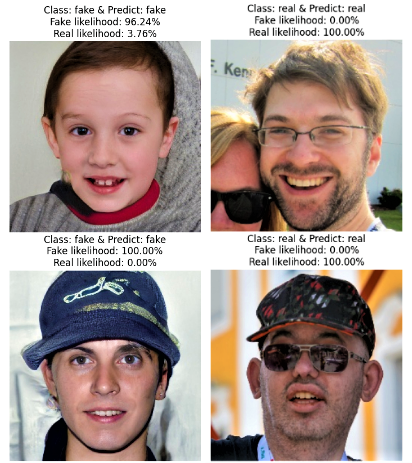}
    \caption{Accurate Identification of employed VFDNET model}
    \label{fig:realfakevit}
\end{figure}

\begin{table}[ht]
    \centering
    \caption{Performance Metrics of the Proposed and Baseline Models}
    \label{tab:model_performance}
    \renewcommand{\arraystretch}{1.2}
    \begin{tabular}{lcccc}
        \toprule
        \textbf{Proposed} & \textbf{Acc.(\%)} & \textbf{Precision(\%)} & \textbf{Recall(\%)} & \textbf{F1-Score(\%)} \\
        \midrule
        MobileNetV3 & 98.00 & 98.11 & 98.09 & 98.09 \\
        VFDNET & 99.13 & 99.00 & 99.00 & 99.00 \\
        ResNet50 & 84.28 & 84.00 & 84.00 & 84.00 \\
        DFCNET  & 95.76 & 92.00 & 91.00 & 89.00 \\
        \bottomrule
    \end{tabular}
\end{table}
\begin{figure}[h]
\centering
\includegraphics[width=\linewidth,height=5cm]{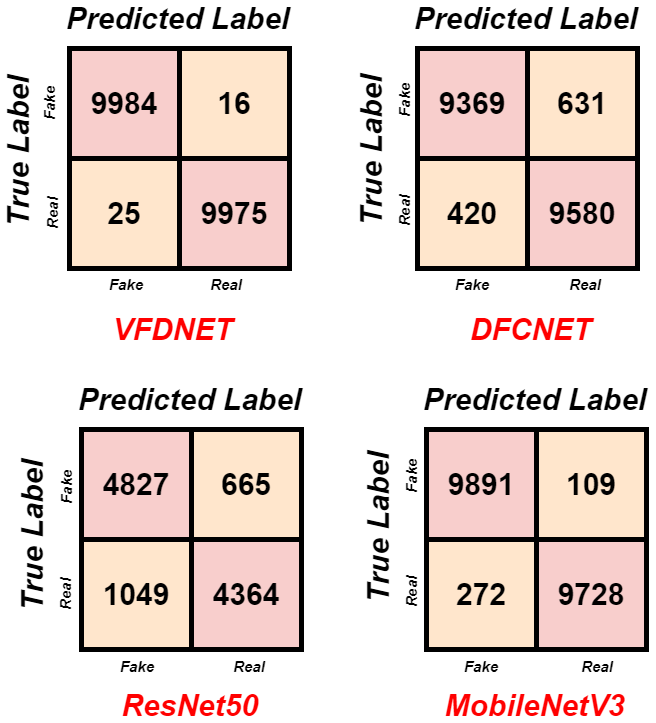}
 \caption{Class-wise Prediction Performance via Confusion Matrices of Proposed Models}
\label{fig:conf.}
\end{figure}
\vspace{1mm} Figure \ref{fig:conf.} represents the confusion matrices of each model tested: VFDNET, DFCNET, ResNet50, and MobileNetV3. VFDNET achieves the highest accuracy levels, correctly labeling 9984 fake and 9975 real images, with only 41 misclassifications in total. DFCNET identified 9369 fake and 9580 real images correctly but exhibited a higher degree of confusion and misclassification of approximately 1051 samples. ResNet50 ranked lowest in weakness was its performance: only 4827 fake and 4364 real images might rightfully be called classification, with a total number of 1714 misclassifications, all indicating poor generalization potential. MobileNetV3 performed well, correctly labeling 9891 fake and 9728 real images, with 381 misclassifications, ranking second after VFDNET. These results therefore indicate VFDNET to be the most potent deepfake detector, followed by MobileNetV3, which is also highly trusted. Figure \ref{fig:realfakevit} illustrates the predictive performance of our best proposed model, VFDNET, on the test dataset images. It displays the predicted class (fake or real) and the corresponding likelihood percentages for both real and counterfeit classifications.
\subsection{Discussion and Comparative Study}
As shown in Table \ref{tab:comparison_existing_methods}, the proposed VFDNET model achieved the highest accuracy 99.13\%, surpassing existing methods such as VGG16 99\% and Ensemble Learning 97.04\%. MobileNetV3 also performed strongly with 98\% accuracy, demonstrating high efficiency. Although DFCNET recorded a balanced accuracy of 95.76\%, it remains comparable to traditional models. Overall, the proposed methods exhibit superior performance, validating their effectiveness for deepfake detection.
\begin{table}[htbp]
\caption{Comparative Analysis of Existing Works and Proposed Methods}
\label{tab:comparison_existing_methods}
\centering
\renewcommand{\arraystretch}{1.2}
\begin{tabular}{@{}l l l c@{}}
\toprule
\textbf{Work} & \textbf{Method} & \textbf{Dataset} & \textbf{Acc. (\%)} \\
\midrule
\cite{zhang2022heterogeneous} & Ensemble Learning & CelebA & 97.04 \\
\cite{rafique2023deep} & ResNet18 & Real and Fake Face & 89.50 \\
\multirow{2}{*}{\cite{article}} 
    & VGG16 & 140k Real and Fake Faces & 99.00 \\
    & DenseNet-121 & 140k Real and Fake Faces & 92.00 \\
\cite{10646310} & ViT & Deepfake and Real Images & 89.91 \\
\midrule
\multirow{3}{*}{\textbf{Proposed}} 
    & DFCNET &  140k Real and Fake Faces & \textbf{95.76} \\
    & VFDNET & 140k Real and Fake Faces & \textbf{99.13} \\
    & MobileNetV3 & 140k Real and Fake Faces & \textbf{98.00} \\
    & ResNet50 & 140k Real and Fake Faces & \textbf{84.28} \\
\bottomrule
\end{tabular}
\end{table}
\section{Conclusion}
The research examined CNN and Vision Transformer systems to detect deepfakes and determined that VFDNET achieved the highest overall performance. Model robustness and generalization improved through the use of preprocessing techniques and augmentation strategies. The research proposes a dependable method for protecting against deepfake threats in actual situations.
 \bibliographystyle{ieeetr}
 \bibliography{ref}
\end{document}